\documentclass{article} 
\usepackage{iclr2016_conference,times}
\usepackage{hyperref}
\usepackage{url}
\usepackage{graphicx}
\usepackage[cmex10]{amsmath}
\usepackage{array}
\usepackage{mdwmath}
\usepackage{mdwtab}
\usepackage{eqparbox}
\usepackage{tikz}
\usepackage{tikz-qtree}
\usepackage{stfloats}
\usepackage{times}
\usepackage{xcolor}
\usepackage{amssymb}
\usepackage{epsfig}
\usepackage{amssymb}
\usepackage[]{algorithm2e}
\usepackage{floatrow}

\newcommand{\argmax}{\arg\!\max}
\definecolor{green}{rgb}{0.2,0.8,0.2}
\definecolor{blue}{rgb}{0.2,0.2,0.8}
\usepackage{xspace}

\newcommand*{\aka}{a.k.a.\@\xspace}
\makeatletter
\newcommand*{\etc}{%
    \@ifnextchar{.}%
        {etc}%
        {etc.\@\xspace}%
}

\title{Sequence Level Training with\\Recurrent Neural Networks}

\author{Marc'Aurelio Ranzato, Sumit Chopra, Michael Auli, Wojciech Zaremba \\
Facebook AI Research\\
\texttt{\{ranzato, spchopra, michealauli, wojciech\}@fb.com}
}

\newcommand{\bh}{\mathbf{h}}

\newcommand{\bx}{\mathbf{x}}
\newcommand{\bc}{\mathbf{c}}
\newcommand{\bone}{\mathbf{1}}

\newcommand{\bs}{\mathbf{s}}

\newcommand{\bo}{\mathbf{o}}
\newcommand{\setr}{\mathcal{R}}

\iclrfinalcopy 

\begin{document}
\maketitle


\begin{abstract}
Many natural language processing applications use language models to generate text.
These models are typically trained to predict the next word in a sequence, given the previous words and some context such as an image. 
However, at test time the model is expected to generate the entire sequence from scratch. 
This discrepancy makes generation brittle, as errors may accumulate along the way. 
We address this issue by proposing a novel sequence level training algorithm that directly optimizes the metric used at test time, such as BLEU or ROUGE.
On three different tasks, our approach outperforms several strong baselines for greedy generation. The method is also competitive when these baselines employ beam search,
 while being several times faster. 
\end{abstract}

\section{Introduction}
Natural language is the most natural form of communication for humans. It is therefore essential that interactive AI systems are capable of generating text~\citep{textgen}.
A wide variety of applications rely on text generation, including machine translation, video/text summarization, question answering, among others. 
From a machine learning perspective, text generation is the problem of predicting a syntactically and semantically correct sequence of consecutive words given some context. For instance, given an image,  generate an appropriate caption or given a sentence in English language, translate it into French. 

Popular choices for text generation models are language models based on n-grams~\citep{kneser+ney1995}, feed-forward neural networks~\citep{nlm}, and recurrent neural networks (RNNs; Mikolov et al., 2010)\nocite{mikolov-2010}. These models when used as is to generate text suffer from two major drawbacks. First, they are trained to predict the next word given the previous ground truth words as input. 
However, at test time, the resulting models are used to generate an entire sequence by predicting one word at a time, and by feeding the generated word back as input at the next time step.
This process is very brittle because the model was trained on a different distribution of inputs, namely, words drawn from the data distribution, as opposed to words drawn from the model distribution. As a result the errors made along the way will quickly accumulate.
We refer to this discrepancy as \textit{exposure bias} which occurs when a model is only exposed to the training data distribution, instead of its own predictions. 
Second, the loss function used to train these models is at the word level. A popular choice is the cross-entropy loss used to maximize the probability of the next correct word. However, the performance of these models is typically evaluated using discrete metrics. One such metric is called BLEU~\citep{bleu} for instance, which measures the n-gram overlap between the model generation and the reference text. Training these models to directly optimize metrics like BLEU is hard because a) these are not differentiable \citep{rosti2011}, and b) combinatorial optimization is required to determine which sub-string maximizes them given some context. Prior attempts~\citep{mcallister2010,he12} at optimizing test metrics were restricted to linear models, or required a large number of samples to work well~\citep{auli2014}.

This paper proposes a novel training algorithm which results in improved text generation compared to standard models. The algorithm addresses the two issues discussed above as follows. First, while training the generative model we avoid the exposure bias by using model predictions at training time. Second, we directly optimize for our final evaluation metric. Our proposed methodology borrows ideas from the reinforcement learning literature~\citep{sutton-rl}. In particular, we build on the REINFORCE algorithm proposed by~\citet{reinforce}, to achieve the above two objectives. While sampling from the model during training is quite a natural step for the REINFORCE algorithm, optimizing directly for any test metric can also be achieved by it. REINFORCE side steps the issues associated with the discrete nature of the optimization by not requiring rewards (or losses) to be differentiable. 
While REINFORCE appears to be well suited to tackle the text generation problem, it suffers from a significant issue. 
The problem setting of text generation has a very large action space which makes it extremely difficult to learn with an initial random policy.
Specifically, the search space for text generation is of size $\mathcal{O}(\mathcal{W}^T)$, where  $\mathcal{W}$ is the number of words in the vocabulary (typically around $10^4$ or more) and $T$ is the length of the sentence (typically around $10$ to $30$). 

Towards that end, we introduce Mixed Incremental Cross-Entropy Reinforce (MIXER), which is our first major contribution of this work. MIXER is an easy-to-implement recipe to make REINFORCE work well for text generation applications. It is based on two key ideas: 
incremental learning and the use of a hybrid loss function which
combines both REINFORCE and cross-entropy (see Sec.~\ref{model-mixer} for details). Both ingredients are essential to training with large action spaces.
In MIXER, the model starts from the optimal policy given by cross-entropy training (as opposed to a random one), from which it then slowly deviates, 
in order to make use of its own predictions, as is done at test time.

Our second contribution is a thorough empirical evaluation on three 
different tasks, namely, Text Summarization, Machine Translation and Image Captioning.
We compare against several strong baselines, including, RNNs trained with cross-entropy and Data as Demonstrator (DAD) \citep{sbengio-nips2015, dad}. 
We also compare MIXER with another simple yet novel model that we propose in this paper. We call it the End-to-End BackProp model (see Sec.~\ref{model-e2e} for details). 
Our results show that MIXER with a simple greedy search achieves much better accuracy compared to the baselines on all the three tasks. In addition we show that MIXER with greedy search is even more accurate than the cross entropy model augmented with beam search at inference time as a post-processing step. This is particularly remarkable because MIXER with greedy search is at least $10$ times faster than the cross entropy model with a beam of size $10$. Lastly, we  note that MIXER and beam search are complementary to each other and can be combined to further improve performance, although the extent of the improvement is task dependent.~\footnote{Code available at: \url{https://github.com/facebookresearch/MIXER}}

\section{Related Work}
Sequence models are typically trained to predict the next
word using the cross-entropy loss. At test time, it is common to use beam search to 
explore  multiple alternative paths~\citep{sutskever2014,bahdanau-iclr2015,rush-2015}.
While this improves generation by typically one or two BLEU points~\citep{bleu}, 
it makes the generation at least $k$ times slower, where $k$ is the number of active 
paths in the beam (see Sec.~\ref{model-xent} for more details).

The idea of improving generation by letting the model use its own predictions at training 
time (the key proposal of this work) was first advocated by~\citet{searn}. In their seminal 
work, the authors first noticed that structured prediction problems
can be cast as a particular instance of reinforcement learning. They then
proposed SEARN, an algorithm to learn such structured
prediction tasks. The basic idea is to let the model use its own
predictions at training time to produce a sequence of actions (e.g.,
the choice of the next word). Then, a search algorithm is
run to determine the optimal action at each time step, and a
classifier (\aka policy) is trained to predict that action. A similar idea was later
proposed by~\citet{dagger} in an imitation learning framework. 
Unfortunately, for text generation it is generally intractable to compute
an oracle of the optimal target word given the words predicted so far.
The oracle issue was later addressed by an algorithm called 
Data As Demonstrator (DAD)~\citep{dad} and applied for text generation by 
\cite{sbengio-nips2015}, whereby the target action at step $k$ is the
$k$-th action taken by the optimal policy (ground truth sequence) regardless of which
input is fed to the system, whether it is ground truth, or the model's
prediction. While DAD usually improves generation, it seems unsatisfactory to force 
the model to predict a certain word regardless of the preceding words 
(see sec.~\ref{model-dad} for more details). 

Finally, REINFORCE has already been used for other applications, such as in 
computer vision~\citep{vmnih-nips2014,xu-icml2015,ba_iclr15}, and for speech recognition~\cite{graves_icml14}. 
While they simply pre-trained with cross-entropy loss, we found that the use of a mixed loss and a more gentle incremental learning scheduling to be important for all the tasks we considered.
\section{Models} \label{model}
\begin{table}[t]
	\footnotesize
    \caption{Text generation models can be described across three dimensions: whether they suffer from exposure bias, 
whether they are trained in an end-to-end manner using back-propagation, 
and whether they are trained to predict one word ahead or the whole sequence.}
    \begin{tabular}{l || c | c | c |c}
      \multicolumn{1}{l||}{\em PROPERTY}  & 
      \multicolumn{1}{l|}{XENT} &
      \multicolumn{1}{l|}{DAD} & \multicolumn{1}{c|}{E2E} & \multicolumn{1}{c}{MIXER}\\
      \hline
      \hline
      {\em avoids exposure bias} & No & Yes & Yes & Yes \\
      \hline
      {\em end-to-end} & No & No & Yes & Yes \\
      \hline
      {\em sequence level} & No & No & No & Yes \\
    \end{tabular}
\label{tab:model_comparison}
\end{table}

The learning algorithms we describe in the following sections are agnostic to
the choice of the underlying model, as long as it is parametric. 
In this work, we focus on Recurrent Neural Networks (RNNs) as they are a popular choice for text generation. In particular, we use standard Elman RNNs~\citep{elman1990} and LSTMs~\citep{lstm}. For the sake of simplicity but without loss of generality, we discuss next Elman RNNs. This is a parametric 
model that at each time step $t$, takes as input a word $w_t \in \mathcal{W}$ as its input, together 
with an internal representation $\bh_t$. $\mathcal{W}$ is the the vocabulary of input words. 
This internal representation $\bh_t$ is a real-valued vector which encodes the history of 
words the model has seen so far. 
Optionally, the RNN can also take as input an  additional context vector $\bc_t$, which 
encodes the context to be used while generating the output. 
In our experiments $\bc_t$ is computed using an attentive decoder 
inspired by \cite{bahdanau-iclr2015} and \citet{rush-2015}, the details of which 
are given in Section ~\ref{sup-material:encoder} of the supplementary material. 
The RNN learns a recursive function to compute $\bh_t$ and 
outputs the distribution over the next word:
\begin{align}
    \bh_{t + 1} &= \phi_\theta(w_t, \bh_t, \bc_t),  \\
	w_{t+1} &\sim p_{\theta}(w |  w_t, \bh_{t+1}) =  p_{\theta}(w |  w_t, \phi_\theta(w_t, \bh_t, \bc_t)).
\end{align}
The parametric expression for $p_\theta$ and $\phi_\theta$ depends 
on the type of RNN. For Elman RNNs we have:
\begin{align}
\label{eq:elman-rnn}
    \bh_{t + 1} &= \sigma(M_i \bone(w_t) + M_h \bh_t + M_c \bc_t), \\
    \bo_{t+1} &= M_o \bh_{t+1}, \label{eq:softmax_input} \\
	w_{t+1} &\sim \mbox{softmax}(\bo_{t+1}),
    \label{eq:xent-distr}
\end{align}
where the parameters of the model $\theta$ are the set of matrices $\{M_o, M_i, M_h, M_c\}$ 
and also the additional parameters used to compute $\bc_t$. $\mbox{Softmax}(\bx)$ is a vector whose components are $e^{x_j} / \sum_k{e^{x_k}}$, and $\bone(i)$ is an indicator vector with only the $i$-th component set to $1$ and the rest to $0$. We assume the first word of the sequence is a special token indicating the beginning of a sequence, denoted by $w_1 = \varnothing$. All entries of the first hidden state $\bh_1$ are set to a constant value.

Next, we are going to introduce both baselines and the model we propose. As we describe these models, it is useful to keep in mind the key characteristics of a text generation system, as outlined in Table~\ref{tab:model_comparison}. There are three dimensions which are important when training a model for text generation: the exposure bias which can adversely affect generation at test time, the ability to fully back-propagate gradients (including with respect to the chosen inputs at each time step), and a loss operating at the sequence level. 
 We will start discussing models that do not possess any of these desirable features, and then move towards models that better satisfy our requirements. The last model we propose, dubbed MIXER, has all the desiderata.

\subsection{Word-Level Training}
We now review a collection of methodologies used for training text generation models which 
optimize the prediction of only one word ahead of time. 
We start with the simplest and the most popular method which optimizes the cross-entropy 
loss at every time step. We then discuss a recently proposed modification to it 
which explicitly uses the model predictions during training. 
We finish by proposing a simple yet novel baseline which uses its model prediction during 
training and also has the ability to back propagate the gradients through the entire sequence. 
While these extensions tend to make generation more robust, 
they still lack explicit supervision at the sequence level. 

\subsubsection{Cross Entropy Training (XENT)} \label{model-xent}
\begin{figure}[!t]
	   \includegraphics[width=0.65\linewidth]{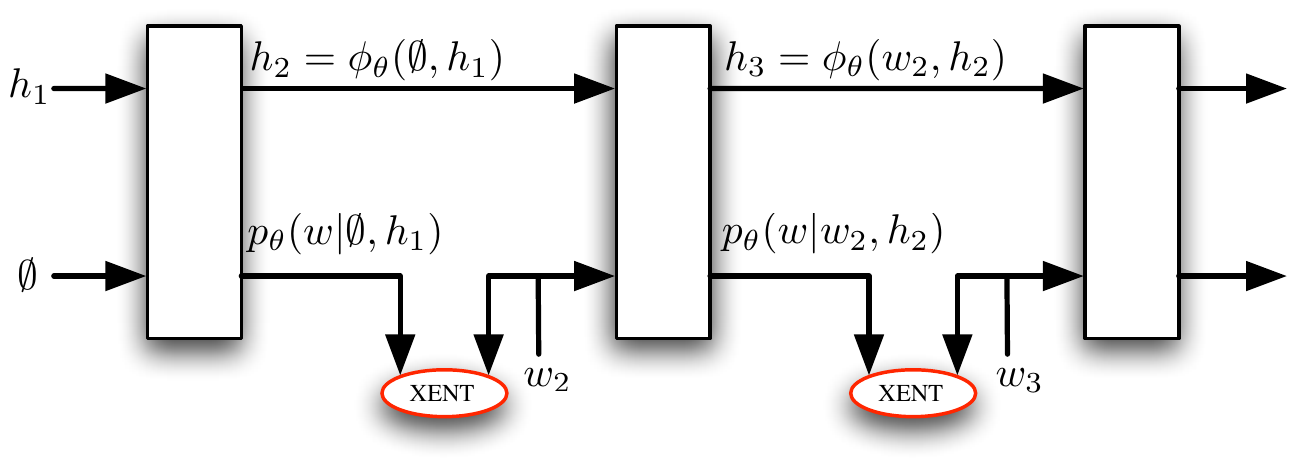}\\
	   \includegraphics[width=0.65\linewidth]{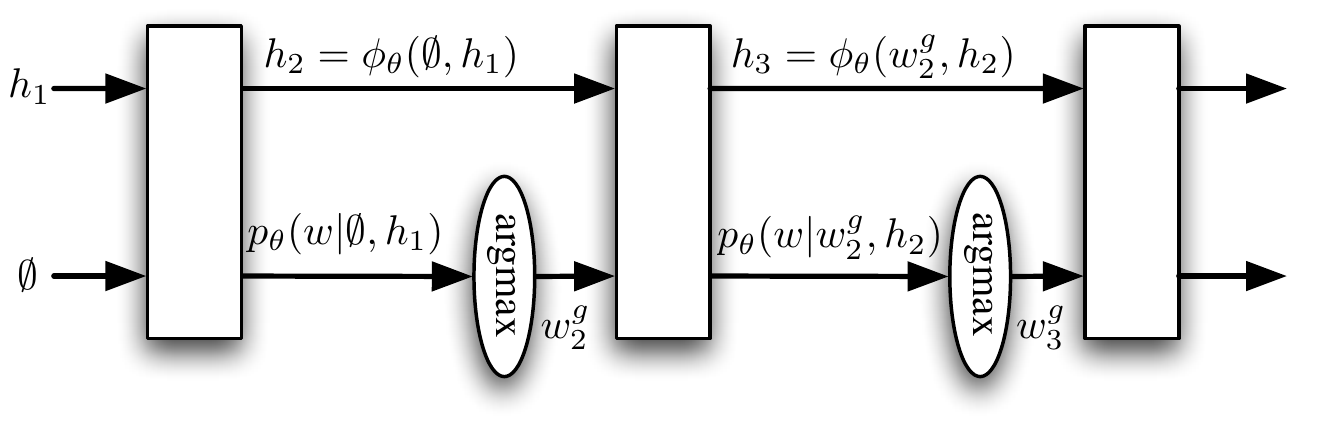}
  \caption{RNN training using XENT (top), and how it is used at test time for generation (bottom). The RNN is unfolded for three time steps in this example. The red oval is a module computing a loss, while the rectangles represent the computation done by the RNN at one step. At the first step, all inputs are given. In the remaining steps, the input words are clamped to ground truth at training time, while they are clamped to model predictions (denoted by $w^g_t$) at test time. Predictions are produced by either taking the argmax or by sampling from the distribution over words. 
  }
  \label{fig:xent}
\end{figure}
Cross-entropy loss (XENT) maximizes the probability of the observed sequence according to the model.
If the target sequence is $[w_1, w_2, \dots, w_T]$, then XENT training involves minimizing: 
\begin{align}
L =	- \log p(w_1, \ldots, w_T) = - \log \prod_{t=1}^T p(w_t | w_1, \ldots, w_{t-1}) = - \sum_{t=1}^T \log p(w_t | w_1, \ldots, w_{t-1}).  \label{eq:xenty}
\end{align}
When using an RNN, each term $p(w_t | w_1, \ldots, w_{t-1})$ is modeled as a parametric function as given in Equation~\eqref{eq:xent-distr}. This loss function trains the model to be good at greedily predicting the next word at each time step without considering the whole sequence. Training proceeds by truncated back-propagation through time~\citep{bptt} with gradient clipping~\citep{mikolov-2010}.

Once trained, one can use the model to generate an entire sequence as follows. Let $w^g_t$ denote the word generated by the model at the $t$-th time step. Then the next word  is generated by: 
\begin{align}
  \label{eq:greedy_gen}
  w^g_{t + 1} = \argmax_w p_{\theta}(w | w^g_t, \bh_{t+1}).
\end{align}
Notice that, the model is trained
to maximize $p_{\theta}(w | w_t, \bh_{t+1})$, where $w_t$ is the word in the ground truth sequence. However, during generation the model is used as  
$p_{\theta}(w | w^g_t, \bh_{t+1})$. In other words, during training the model is only exposed 
to the ground truth words. However, at test time the model has only
access to its own predictions, which may not be correct. As a result, during generation the model can potentially deviate quite far from the actual sequence to be generated. Figure~\ref{fig:xent} illustrates this discrepancy. 

The generation described by Eq.~\eqref{eq:greedy_gen} is
a greedy left-to-right process which does not necessarily produce
the most likely sequence according to the model, because:
\begin{align*}
    \prod_{t=1}^T ~\max_{w_{t+1}}~p_{\theta}(w_{t+1} | w^g_t, \bh_{t+1}) \leq
    \max_{w_1, \dots, w_T}~\prod_{t=1}^Tp_{\theta}(w_{t+1} | w^g_t, \bh_{t+1})
\end{align*}
The most likely sequence $[w_1, w_2, \dots, w_T]$ might contain a word  $w_t$ which is sub-optimal at an intermediate time-step $t$. This phenomena is commonly known as a {\it search error}. 
One popular way to reduce the effect of search error is to pursue not only one but $k$ next word 
candidates at each point. While still approximate, this strategy can recover 
higher scoring sequences that are often also better in terms of our final evaluation metric.
This process is commonly know as {\it Beam Search}. The downside of using beam search is that it significantly slows down 
the generation process. The time complexity 
grows linearly in the number of beams $k$, because we need to perform 
$k$ forward passes for our network, which is the most time intensive operation. 
The details of the 
Beam Search algorithm are described in Section~\ref{sup-material:beam_search}.


\subsubsection{Data As Demonstrator (DAD)} \label{model-dad}
\begin{figure}[!t]
\begin{center}
 \includegraphics[width=0.7\linewidth]{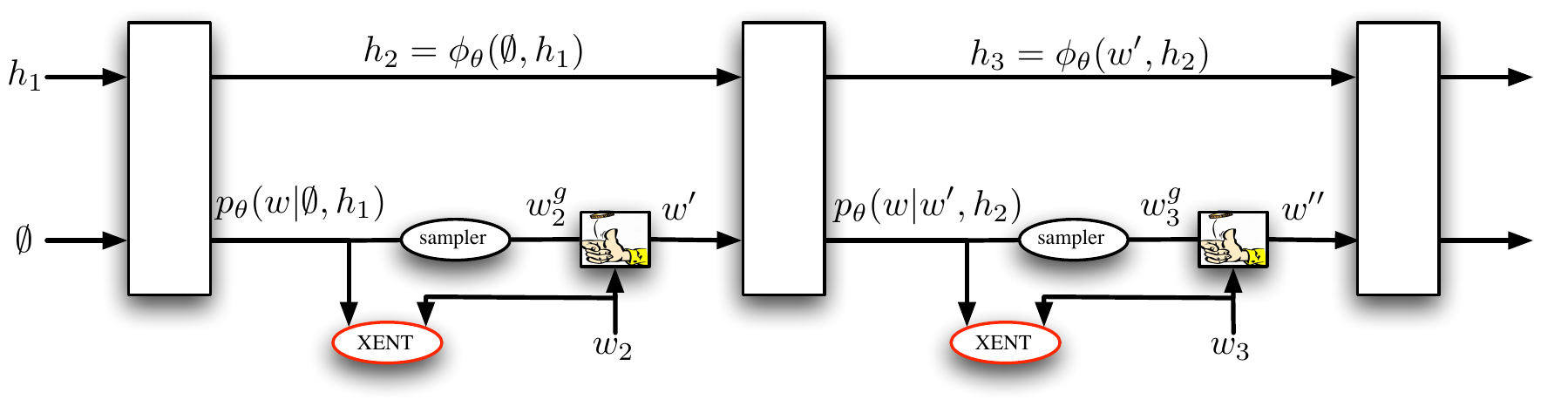}
\end{center}
\caption{Illustration of DAD~\citep{sbengio-nips2015,
    dad}. Training
  proceeds similar to XENT, except that at each time step we choose with
  a certain probability whether to take the previous model prediction or
  the ground truth word. Notice how a) gradients are not
  back-propagated through the eventual model predictions $w^g_t$, and
  b) the XENT loss always uses as target the next word in the reference
  sequence, even when the input is $w^g_t$.}
\label{fig:dad}
\end{figure}
Conventional training with XENT suffers from exposure bias since  
training uses ground truth words as opposed to model predictions.
DAD, proposed in~\citep{dad} and also used in~\citep{sbengio-nips2015} for sequence generation, addresses this issue by mixing the ground truth training data with model predictions.
At each time step and with a certain probability, DAD takes as input either the prediction from the model at the previous time step or the ground truth data. \citet{sbengio-nips2015} proposed  different 
annealing schedules for the probability of choosing the ground truth word. The annealing schedules are such that at the beginning, the algorithm always chooses the ground truth words. However, as the training progresses the model predictions are selected more often. 
This has the effect of making the model somewhat more aware of how it will be used at test time. Figure~\ref{fig:dad} illustrates the algorithm. 

A major limitation of DAD is that at every time step the target labels are always selected from the ground truth data,  regardless of how the input was chosen. As a result, the targets may not be aligned with the generated sequence, forcing the model to predict a potentially incorrect sequence. 
For instance, if the ground truth sequence is ``I took a long
walk" and the model has so far predicted ``I took a walk", DAD will force the model to predict the word ``walk" a second time. 
Finally, gradients are not back-propagated through the samples drawn by the model and the XENT loss is still at the word level. 
It is not well understood how these problems affect generation.

\subsubsection{End-to-End BackProp (E2E)} 
\label{model-e2e}
\begin{figure}[!t]
\begin{center}
 \includegraphics[width=0.7\linewidth]{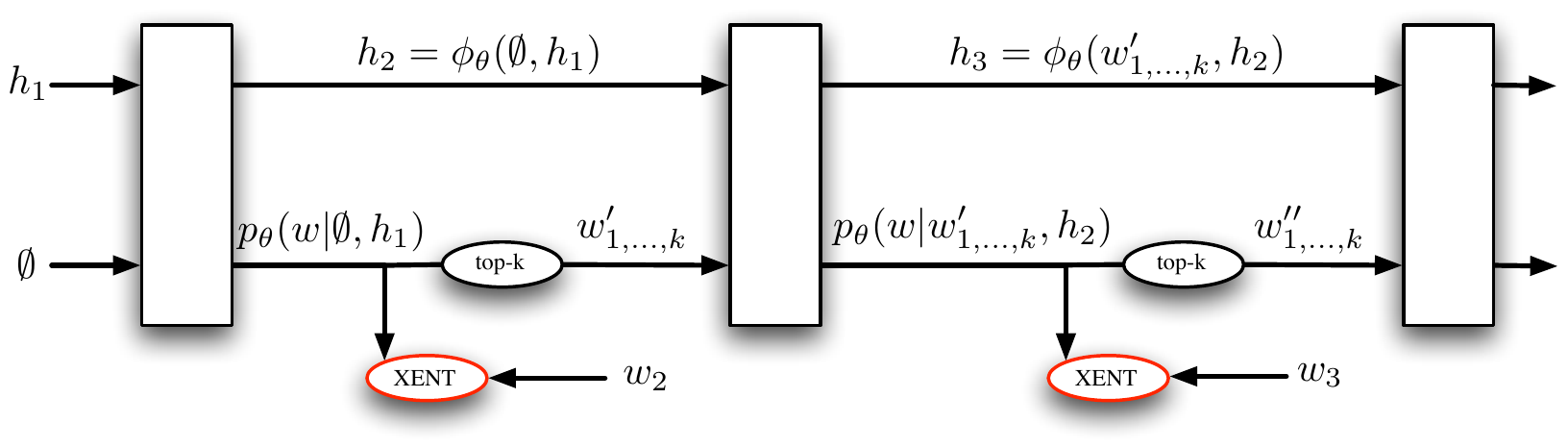}
\end{center}
\caption{Illustration of the End-to-End BackProp method. The first 
  steps of the unrolled sequence (here just the first step) are exactly the same
  as in a regular RNN trained with cross-entropy. However, in the remaining
  steps the input to each module is a sparse
  vector whose non-zero entries are the $k$ largest probabilities
  of the distribution predicted at the previous time step. Errors are
  back-propagated through these inputs as well.}
\label{fig:e2e}
\end{figure}

The novel E2E algorithm is perhaps the most natural and na\"{\i}ve approach approximating sequence level training, which can also be interpreted as a computationally efficient approximation to beam search. 
The key idea is that at time step $t + 1$ we propagate as input the top $k$ words predicted at the previous time step instead of the ground truth word. 
Specifically, we take the output distribution over words from the previous time step $t$, and pass it through a $k$-max layer. 
This layer zeros all but the $k$ largest values and re-normalizes 
them to sum to one. We thus have: 
\begin{equation}
\{i_{t+1,j}, v_{t+1,j}\}_{j=1, \dots, k} = \mbox{k-max } p_\theta(w_{t+1} |  w_t, h_t), \label{eq:e2e}
\end{equation}
where $i_{t+1,j}$ are indexes of the words with $k$ largest probabilities and $v_{t+1,j}$ are their corresponding scores. 
At the time step $t+1$, we take the $k$ largest scoring previous words as input whose 
contributions is weighted by their scores $v$'s. 
Smoothing the input this way makes the whole process 
differentiable and trainable using standard back-propagation.
Compared to beam search, this can be interpreted as fusing the $k$ possible next 
hypotheses together into a single path, as illustrated in Figure~\ref{fig:e2e}. 
In practice we also employ a schedule, whereby we use only the ground truth words 
at the beginning and gradually let the model use its own top-$k$ predictions as training proceeds. 

While this algorithm is a simple way to expose the model to its own predictions, the loss function optimized is still XENT at each time step. 
There is no explicit supervision at the sequence level while training the model. 

\subsection{Sequence Level Training}
We now introduce a novel algorithm for sequence level training, which we call Mixed Incremental Cross-Entropy Reinforce (MIXER). The proposed method avoids the exposure bias problem, 
and also directly optimizes for the final evaluation metric. 
Since MIXER is an extension of the REINFORCE algorithm, we first describe REINFORCE 
from the perspective of sequence generation. 

\subsubsection{REINFORCE} \label{model-reinforce}
In order to apply the REINFORCE algorithm~\citep{reinforce, zaremba-arxiv2015} to the problem of sequence generation we cast our problem in the reinforcement learning (RL) framework~\citep{sutton-rl}. Our generative model (the RNN) can be viewed as an {\em agent}, which interacts with the external environment (the words and the context vector it sees as input at every time step). The parameters of this agent defines a {\em policy}, whose execution results in the agent picking an {\em action}. In the sequence generation setting, an action refers to predicting the next word in the sequence at each time step. 
After taking an action the agent updates its internal state (the hidden units of RNN). Once the agent has reached the end of a sequence, it observes a {\em reward}. We can choose any reward function. Here, we use BLEU~\citep{bleu} and ROUGE-2~\citep{rouge} since these are the metrics we use at test time. BLEU is essentially a geometric mean over n-gram precision scores  as well as a brevity penalty~\citep{liang2006}; in this work, we consider up to $4$-grams. ROUGE-2 is instead recall over bi-grams. 
Like in {\em imitation learning}, we have a training set of optimal sequences of actions. 
During training we choose actions according to the current policy and only observe a reward at the end of the sequence (or after maximum sequence length), by comparing the sequence of actions from the current policy against the optimal action sequence.
The goal of training is to find the parameters of the agent that maximize the expected reward.
We define our loss as the negative expected reward:
\begin{align}
L_{\theta} = - \sum_{w^g_1, \dots, w^g_T} p_\theta(w^g_1, \dots,
w^g_T) r(w^g_1, \dots, w^g_T) = -\mathbb{E}_{[w_1^g, \dots w^g_T] \sim p_\theta} r(w^g_1, \dots, w^g_T), \label{eq:reinforce-loss}
\end{align}
where $w^g_n$ is the word chosen by our model at the $n$-th time step, and $r$ is the reward associated with the generated sequence. 
In practice, we approximate this expectation with a single sample
from the distribution of actions implemented by the RNN (right hand side of the equation above and Figure~\ref{fig:plan} of Supplementary Material). 
We refer the reader to prior work~\citep{zaremba-arxiv2015,reinforce} for the full derivation of the gradients. Here, we directly report the partial derivatives and their interpretation. The derivatives w.r.t. parameters are:
\begin{equation}
\frac{\partial L_{\theta}}{\partial \theta} = \sum_t \frac{\partial L_{\theta}}{\partial \bo_t}
\frac{\partial \bo_t}{\partial \theta} \label{eq:reinf-deriv1}
\end{equation}
where $\bo_t$ is the input to the softmax. 
The gradient of the loss $L_{\theta}$ with respect to $\bo_t$ is given by:
\begin{equation}
\frac{\partial L_{\theta}}{\partial \bo_t} = \left( r(w^g_1, \dots, w^g_T) - \bar{r}_{t+1} \right) \left( p_\theta(w_{t+1} | w^g_{t}, \bh_{t+1}, \bc_t) - \bone(w^g_{t+1}) \right),
 \label{eq:reinf-deriv2}
\end{equation}
where $\bar{r}_{t+1}$ is the average reward at time $t + 1$. 

The interpretation of this weight update rule is straightforward. While
Equation~\ref{eq:reinf-deriv1} is standard back-propagation (a.k.a. chain
rule), Equation~\ref{eq:reinf-deriv2} is almost exactly the same as the gradient of a multi-class logistic regression classifier. In logistic regression, the gradient is the difference between the prediction and the actual 1-of-N representation of the target word:
\begin{equation*}
\frac{\partial L^{\mbox{\small XENT}}_{\theta}}{\partial \bo_t} 
 =  p_\theta(w_{t+1} | w_{t}, \bh_{t+1}, \bc_t) - \bone(w_{t+1})
\end{equation*}
Therefore, Equation~\ref{eq:reinf-deriv2} says that the chosen word $w^g_{t+1}$
acts like a surrogate target for our output distribution,
$p_\theta(w_{t+1}|w^g_{t}, \bh_{t+1}, \bc_t)$ at time $t$. REINFORCE first establishes a baseline $\bar{r}_{t+1}$,
and then either encourages a word choice $w^g_{t+1}$ if $r > \bar{r}_{t+1}$, 
or discourages it if $r < \bar{r}_{t+1}$. The actual derivation suggests that the choice of this average reward $\bar{r}_t$ is useful to decrease the variance of the gradient estimator since in Equation~\ref{eq:reinforce-loss} we use a single sample from the distribution of actions. 

In our implementation, the baseline $\bar{r}_t$ is estimated by a linear regressor which takes as input the hidden states $\bh_t$ of the RNN. The regressor is an unbiased estimator of future rewards since it only uses past information. The parameters of the regressor are
trained by minimizing the mean squared loss: $||\bar{r}_t - r||^2$.
In order to prevent feedback loops, we do not backpropagate this error through
the recurrent network~\citep{zaremba-arxiv2015}. 

REINFORCE is an elegant algorithm to train at the sequence level using {\em any} user-defined reward. In this work, we use BLEU and ROUGE-2 as reward, however one could just as easily use any other metric.
When presented as is, one major drawback associated with the algorithm is that it assumes a random
policy to start with. This assumption can make the learning for large action spaces very challenging.
Unfortunately, text generation is such a setting where the cardinality of the action set is in the order of $10^4$ (the number of words in the vocabulary). 
This leads to a very high branching factor where it is extremely hard for a random policy to improve in any reasonable amount of time. 
In the next section we describe the MIXER algorithm which addresses these issues, better targeting
text generation applications.

\subsubsection{Mixed Incremental Cross-Entropy Reinforce (MIXER)} \label{model-mixer}
The MIXER algorithm borrows ideas both from DAGGER~\citep{dagger} and 
DAD~\citep{dad, sbengio-nips2015} and modifies the REINFORCE appropriately. 
The first key idea is to change the initial policy of REINFORCE to make sure
the model can effectively deal with the large action space of text generation.
Instead of starting from a poor random policy and training the model to converge 
towards the optimal policy, we do the exact opposite. We start from the optimal 
policy and then slowly deviate from it to let the model explore and make use 
of its own predictions.
We first train the RNN with the cross-entropy loss for
$N^{\mbox{\small{XENT}}}$ epochs using the ground truth sequences.
This ensures that we start off with a much better policy than random 
because now the model can focus on a good part of the search space.
This can be better understood by comparing the perplexity of a language model 
that is randomly initialized versus one that is trained. Perplexity is a measure of uncertainty of the prediction and, roughly speaking, it corresponds to the average number of words the model is `hesitating' about when making a prediction. A good language model trained on one of our data sets has perplexity of $50$, whereas a random model is likely to have
perplexity close to the size of the vocabulary, which is about $10,000$.
\begin{figure}[!t]
\begin{center}
 \includegraphics[width=0.75\linewidth]{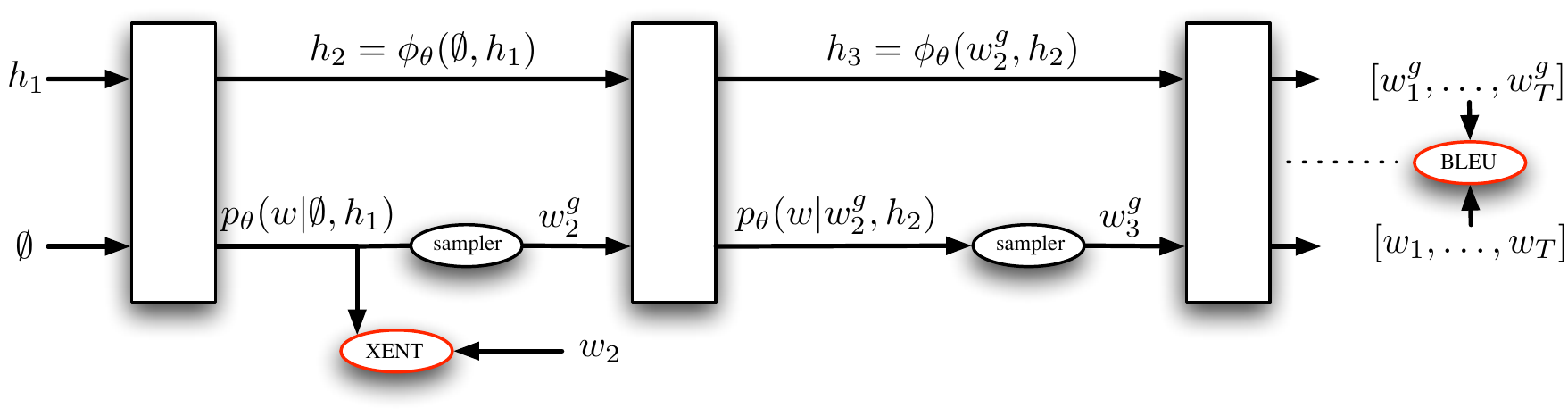}
\end{center}
\caption{Illustration of MIXER. In the first $s$ unrolling steps (here $s=1$),
  the network resembles a standard RNN trained by XENT. In the
  remaining steps, the input to each module is a sample from the
  distribution over words produced at the previous time step. Once the
  end of sentence is reached (or the maximum sequence length), a
  reward is computed, e.g., BLEU. REINFORCE is then
  used to back-propagate the gradients through the sequence of
  samplers. We employ an annealing schedule on $s$, starting with 
  $s$ equal to the maximum sequence length $T$ and finishing with $s = 1$.
 }
\label{fig:mixer}
\end{figure}

The second idea is to introduce model predictions during training 
with an annealing schedule in order to gradually teach the model to produce stable sequences. 
Let $T$ be the length of the sequence.
After the initial $N^{\mbox{\small{XENT}}}$ epochs, we continue training the 
model for $N^{\mbox{\small{XE+R}}}$ epochs, such that, for every sequence 
we use the XENT loss for the first ($T - \Delta$) steps, and the REINFORCE algorithm 
for the remaining $\Delta$ steps.
In our experiments $\Delta$ is typically set to two or three. 
Next we anneal the number of steps for which we use the XENT loss for every sequence to  
($T - 2 \Delta$) and repeat the training for another $N^{\mbox{\small{XE+R}}}$ epochs. 
We repeat this process until only REINFORCE is used to train the whole sequence.
See Algorithm~\ref{alg:mixer} for the pseudo-code. 


We call this algorithm Mixed Incremental Cross-Entropy Reinforce (MIXER)
because we combine both XENT and REINFORCE, and we use incremental 
learning (a.k.a. curriculum learning). 
The overall algorithm is illustrated in Figure~\ref{fig:mixer}. 
By the end of training, the model can make effective use of its own 
predictions in-line with its use at test time.

\begin{algorithm}[t]
\footnotesize
 \KwData{a set of sequences with their corresponding context.}
 \KwResult{RNN optimized for generation.}
 Initialize RNN at random and set $N^{\mbox{\small{XENT}}}$, $N^{\mbox{\small{XE+R}}}$
 and $\Delta$\; 
 \For{$s$ = $T$, $1$, $-\Delta$}{
   \eIf{ $s$ == $T$ }{
     train RNN for $N^{\mbox{\small{XENT}}}$ epochs using XENT only\;
     }{
     train RNN for $N^{\mbox{\small{XE+R}}}$ epochs. Use XENT loss in the
     first $s$ steps, and REINFORCE (sampling from the model) in the remaining $T - s$ steps\;
   }
 }
 \caption{MIXER pseudo-code.}
 \label{alg:mixer}
\end{algorithm}

\section{Experiments} \label{sec:experiments}
In all our experiments, we train conditional RNNs by unfolding them
up to a certain maximum length. 
We chose this length to cover about $95\%$ of the target sentences in the data sets we consider.
The remaining sentences are cropped to the chosen maximum length. 
For training, we use stochastic gradient descent
with mini-batches of size $32$ and we reset the hidden states at the
beginning of each sequence. Before updating the parameters we
re-scale the gradients if their norm is above $10$~\citep{mikolov-2010}.
We search over the values of hyper-parameter, such as the initial learning rate, 
the various scheduling parameters, number of epochs, \etc, using a held-out validation set. 
We then take the model that performed best on the validation set and compute BLEU or ROUGE 
score on the test set. In the following sections we report results on the test set only. 
Greedy generation is performed by taking the most likely word at each time step.~\footnote{Code available at: \url{https://github.com/facebookresearch/MIXER}} 

\subsection{Text Summarization}
We consider the problem of abstractive summarization where, 
given a piece of ``source" text, we aim at generating its summary (the ``target" text)
such that its meaning is intact.  
The data set we use to train and evaluate our models consists of a 
subset of the Gigaword corpus~\citep{gigaword} as described in~\citet{rush-2015}. 
This is a collection of news articles taken from different sources over the past two decades. 
Our version is organized as a set of example pairs, where each pair is composed of the 
first sentence of a news article (the source sentence) and its corresponding headline (the target sentence). 
We pre-process the data in the same way as in~\citep{rush-2015}, which consists of lower-casing 
and replacing the infrequent words with a special token denoted by ``$<$unk$>$". After 
pre-processing there are $12321$ unique words in the source dictionary and $6828$ words in the target dictionary. The number of sample pairs in the training, validation and test set are $179414$, $22568$, 
and $22259$ respectively. The average sequence length of the target headline is about $10$ words. 
We considered sequences up to $15$ words to comply with our initial constraint of covering at least
$95$\% of the data.

Our generative model is a conditional Elman RNN (Equation~\ref{eq:elman-rnn}) with $128$ hidden units, 
where the conditioning vector $\bc_t$  is provided by a convolutional attentive encoder, 
similar to the one described in Section 3.2 of ~\citet{rush-2015} and inspired by 
\cite{bahdanau-iclr2015}. The details of our attentive 
encoder are mentioned in Section~\ref{sup-material:encoder} of the Supplementary Material.
We also tried LSTMs as our generative model for this task, however it did not improve performance. We conjecture this is due to the fact that the target sentences in this data set are rather short.

\subsection{Machine Translation}
For the translation task, our generative model is an LSTM with $256$ hidden units and it uses the same attentive encoder architecture as the one used for summarization. 
We use data from the German-English machine translation track of the
IWSLT 2014 evaluation campaign \citep{cettolo2014}.
The corpus consists of sentence-aligned subtitles of TED and TEDx 
talks. We pre-process the training data using the tokenizer of the Moses 
toolkit \citep{koehn2007} and remove sentences longer than $50$ words as well as casing.  
The training data comprises of about $153000$ sentences where the average English 
sentence is $17.5$ words long and the average German sentence is 
$18.5$ words long. In order to retain at least $95\%$ of this data, we unrolled our RNN for $25$ steps.
Our validation set comprises of $6969$ sentence pairs which was taken from the training data. The test set is a concatenation of dev2010, dev2012, tst2010, tst2011 and tst2012 
which results in $6750$ sentence pairs. The English dictionary has $22822$ words while the German has $32009$ words. 
\subsection{Image Captioning}
For the image captioning task, we use the MSCOCO
dataset~\citep{mscoco}. We use the entire training set provided by the authors, which consists of around $80$k images. We then took the original validation set (consisting of around $40$k images) and randomly sampled (without replacement) $5000$ images for validation and another $5000$ for test. 
There are $5$ different captions for each image. 
At training time we sample one of
these captions, while at test time we report the maximum BLEU score across the five captions.
The context is represented by 1024 features extracted by a Convolutional Neural Network (CNN) trained
on the Imagenet dataset~\citep{imagenet_cvpr09}; we do not back-propagate through these features. We use a similar experimental set up as described in ~\citet{sbengio-nips2015}. The RNN is a single layer LSTM with 
$512$ hidden units and the image features are provided to the generative model as the first word in the sequence. We pre-process the captions by lower-casing all words and replacing all the words which appear less than 3 times with a special token ``$<$unk$>$". As a result the total number of unique words in our dataset is $10012$. Keeping in mind the $95\%$ rule, we unroll the RNN for $15$ steps.

\subsection{Results}
\begin{figure}[!t]
\centering
\begin{minipage}[c][][c]{.4\textwidth}
\centering
\begin{tabular}{l || l | l | l |l}
\multicolumn{1}{c||}{\emph{TASK} }  & 
      \multicolumn{1}{c|}{XENT} &
      \multicolumn{1}{c|}{DAD} & \multicolumn{1}{c|}{E2E} & \multicolumn{1}{c}{MIXER}\\
      \hline
      \hline
      {\em summarization} & 13.01 & 12.18 & 12.78 &  \bf{16.22} \\
      \hline
      {\em translation} & 17.74 & 20.12 & 17.77 & {\bf 20.73} \\
      \hline
      {\em image captioning} & 27.8 & 28.16 & 26.42 & \bf{29.16} \\
    \end{tabular}
\end{minipage}\hfill
\begin{minipage}[c][][c]{.4\textwidth}
\centering
\includegraphics[width=0.8\linewidth]{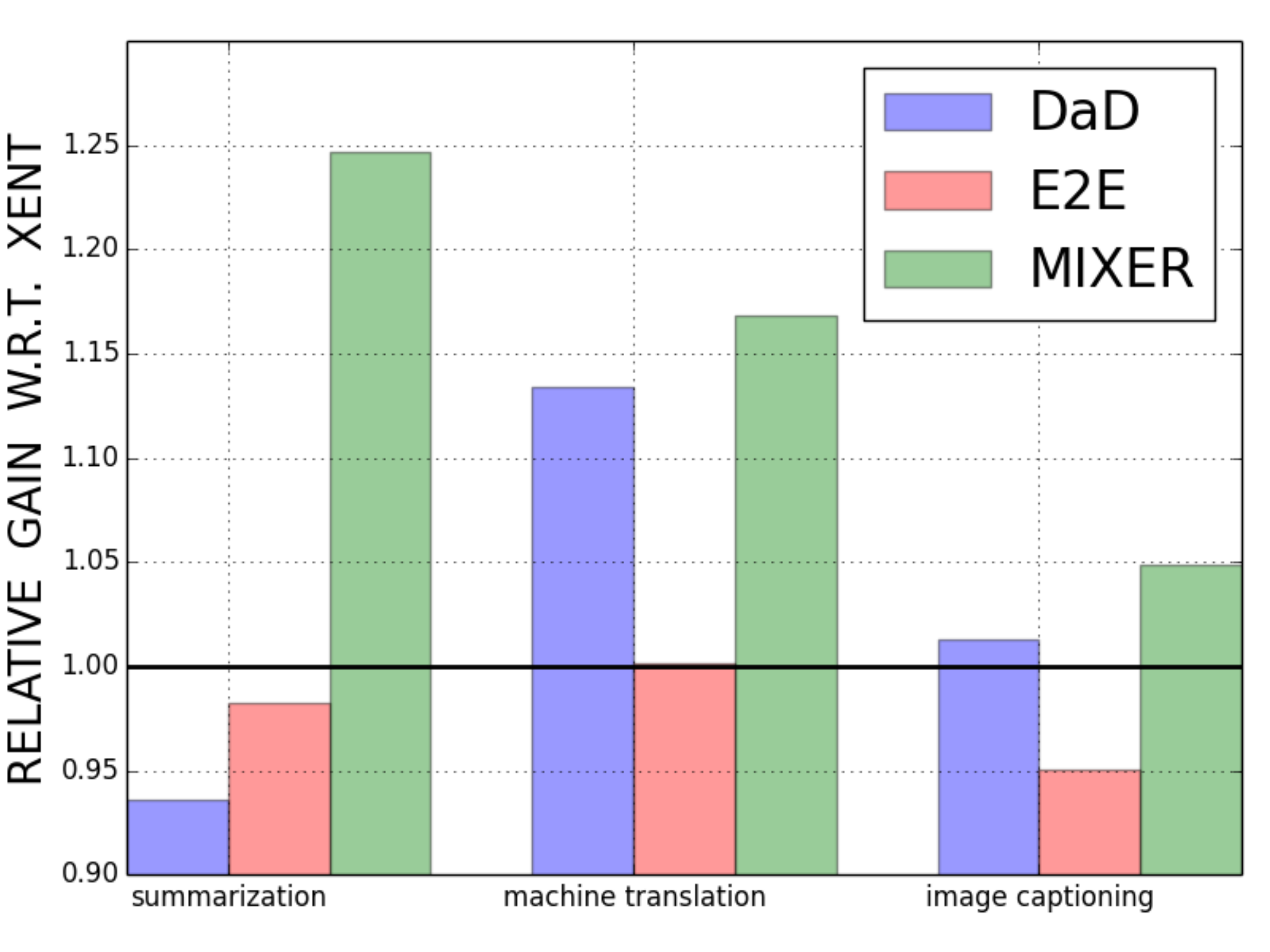}
\vspace{-.30cm}
\caption{Left: BLEU-4 (translation and image captioning) and ROUGE-2 (summarization) scores using greedy generation. Right: Relative gains
  produced by DAD, E2E and MIXER on the three tasks. 
  The relative gain is computed as the ratio between the score of a model over the score of the reference XENT model on the same task. The horizontal line indicates the performance of XENT.}
\label{fig:gain}
\end{minipage}
\end{figure}

In order to validate MIXER, we compute BLEU score on the machine translation and image captioning task, and ROUGE on the summarization task. 
The input provided to the system is only the context and the beginning of sentence token. We apply the same protocol to the baseline methods as well. The scores on the test set are reported in Figure~\ref{fig:gain}. 
\begin{figure}[!t]
\begin{center}
 \includegraphics[width=.35\linewidth]{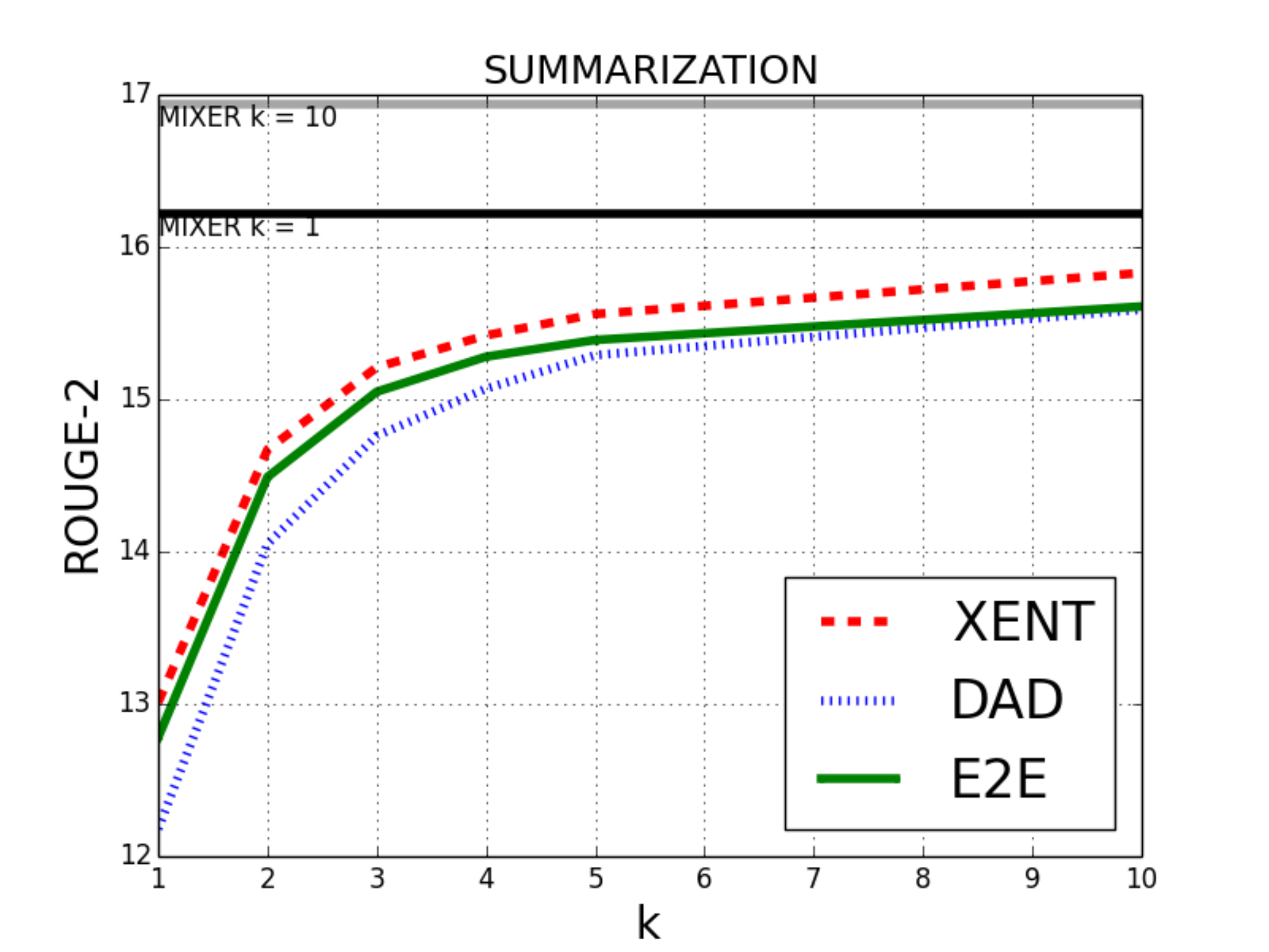}
 \hspace{-.6cm}
 \includegraphics[width=.35\linewidth]{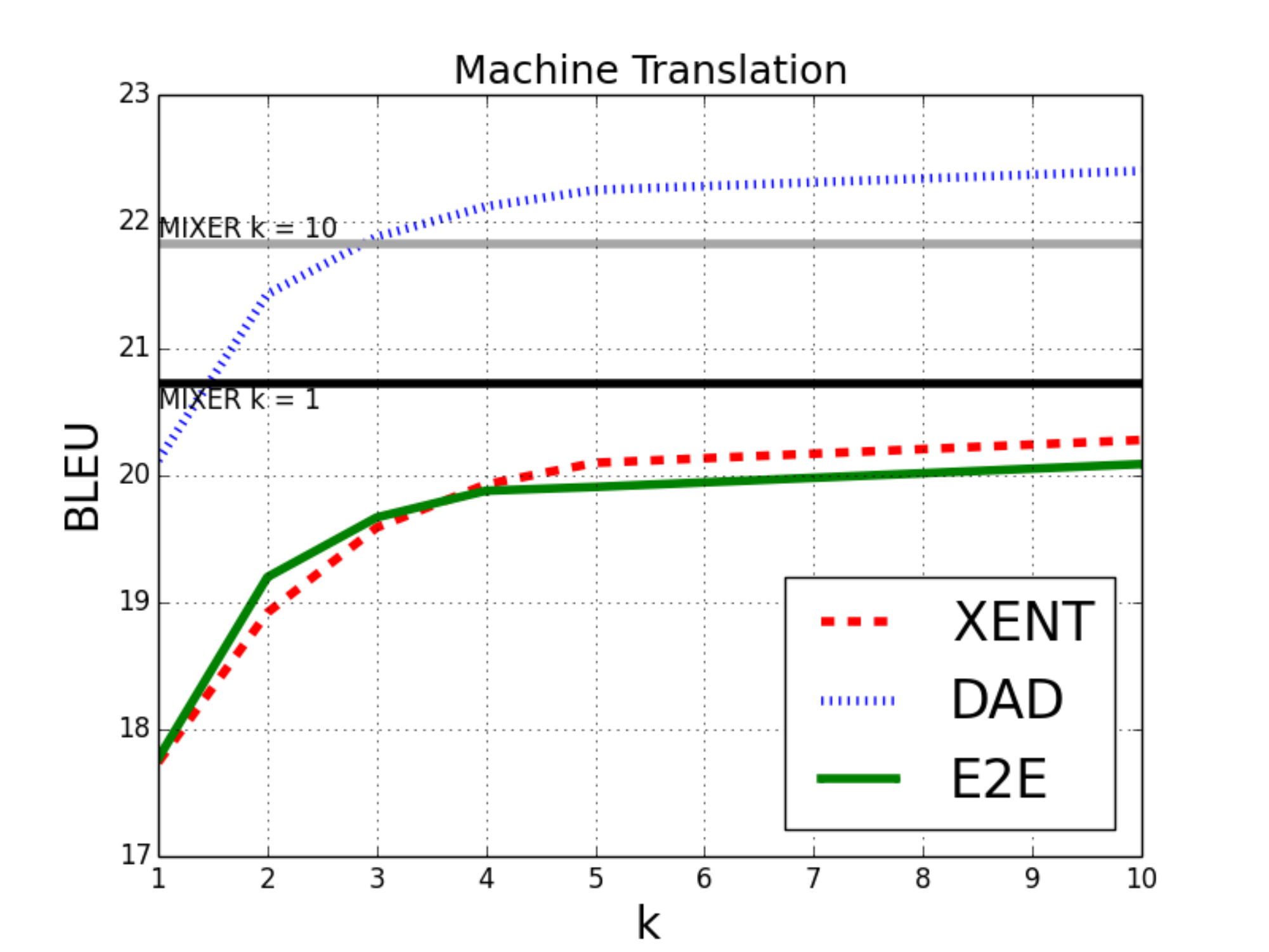}
 \hspace{-.6cm}
  \includegraphics[width=.35\linewidth]{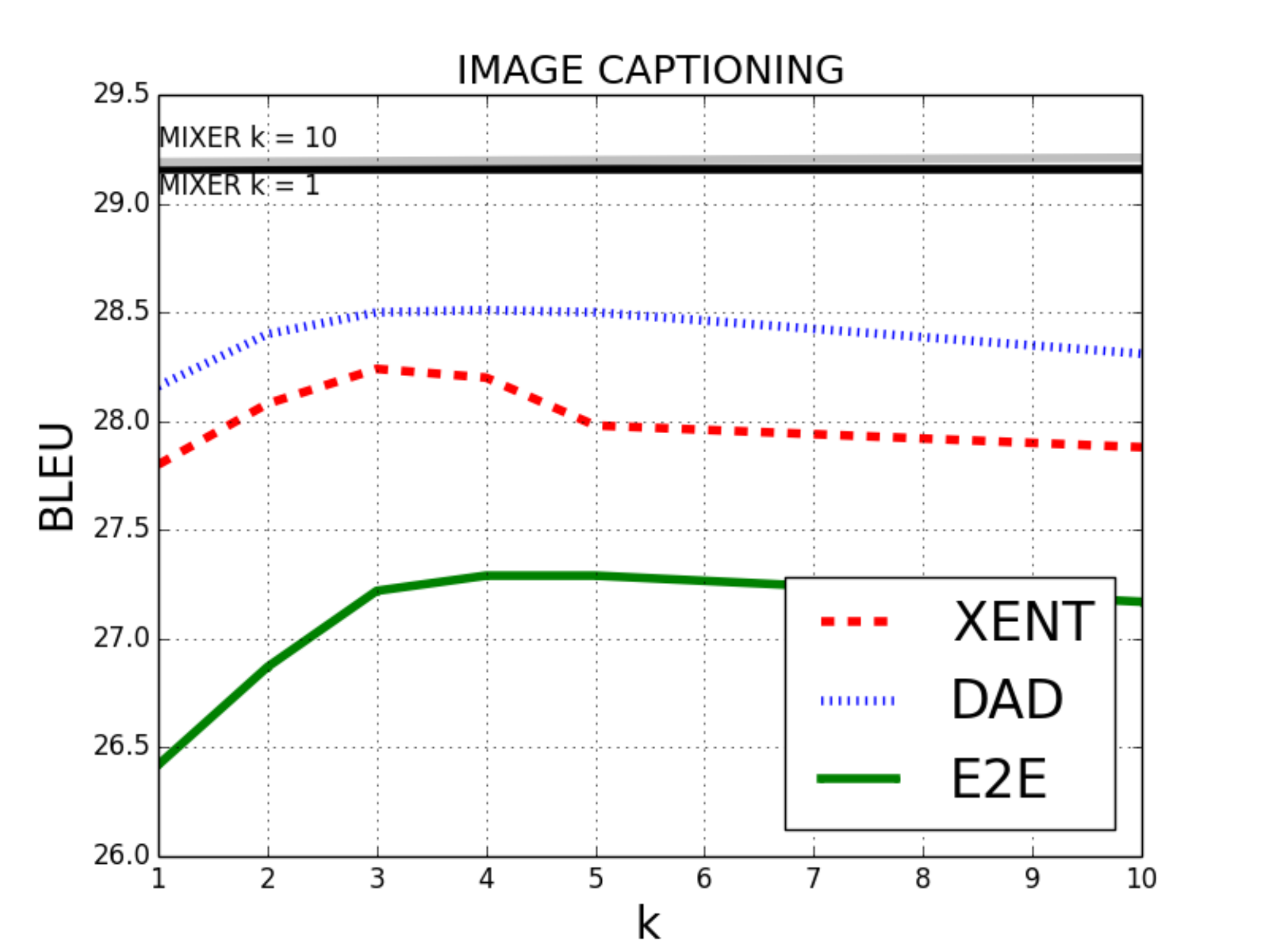}
\end{center}
\vspace{-.2cm}
\caption{Test score (ROUGE for summarization and BLEU for machine translation and image captioning) as a function of the number of hypotheses $k$ in the beam search. Beam search always improves performance, although the amount depends on the task. The dark line shows the performance of MIXER using greedy generation, while the gray line shows MIXER using beam search with $k=10$.}
\label{fig:beam_search}
\end{figure}

We observe that MIXER produces the best generations and improves generation over XENT by $1$ to $3$ points across all the tasks. 
Unfortunately the E2E approach did not prove to be very effective. Training at the sequence level and directly optimizing for testing score yields better generations than turning a sequence of discrete decisions into a differentiable process amenable to standard back-propagation of the error. 
DAD is usually better than the XENT, but not as good as MIXER.  

Overall, these experiments demonstrate the importance of optimizing for the metric used at test time. In summarization for instance, XENT and MIXER trained with ROUGE achieve a poor performance in terms of BLEU (8.16 and 5.80 versus 9.32 of MIXER trained with BLEU); likewise, MIXER trained with BLEU does not achieve as good ROUGE score as 
a MIXER optimizing ROUGE at training time as well (15.1 versus 16.22, see also Figure~\ref{fig:summarization_bleu_rouge} in Supplementary Material).

Next, we experimented with beam search. The results in Figure~\ref{fig:beam_search} suggest that all methods, including MIXER, improve the quality of their generation by using beam search. However, the extent of the improvement is very much task dependent. We observe that the greedy performance of MIXER (i.e., {\em without} beam search) cannot be matched by baselines using beam search in two out of the three tasks. Moreover, MIXER is several times faster since it relies only on greedy search.

It is worth mentioning that the REINFORCE baseline did not work for these applications. Exploration from a random policy has little chance of success. We do not report it since we were never able to make it converge within a reasonable amount of time. Using the hybrid XENT-REINFORCE loss {\em without} incremental learning is also insufficient to make training take off from random chance. In order to gain some insight on what kind of schedule works, we report in Table~\ref{tab:scheduling} of Supplementary Material the best values we found after grid search over the hyper-parameters of MIXER.
Finally, we report some anecdotal examples of MIXER generation in Figure~\ref{fig:generation} of Supplementary Material. 

\section{Conclusions}

Our work is motivated by two major deficiencies in training the current generative models for text generation: exposure bias and a loss which does not operate at the sequence level.
While Reinforcement learning can potentially address these issues, it struggles in settings when 
there are very large action spaces, such as in text generation. Towards that end, 
we propose the MIXER algorithm, which deals with these issues and enables successful training of reinforcement learning models for text generation. 
We achieve this by replacing the initial random policy with the optimal policy of a cross-entropy trained model and by gradually exposing the model more and more to its own predictions in an incremental learning framework.



Our results show that MIXER outperforms three strong baselines for greedy generation and it is very competitive with beam search. 
The approach we propose is agnostic to the underlying model or the form of the reward function. 
In future work we would like to design better estimation techniques for the average reward $\bar{r}_t$, because poor estimates can lead to slow convergence of both REINFORCE and MIXER. 
Finally, our training algorithm relies on a single sample while it would be interesting to investigate the effect of more comprehensive search methods at training time.

\subsubsection*{Acknowledgments}
The authors would like to thank David Grangier, Tomas Mikolov, Leon Bottou, Ronan Collobert and Laurens van der Maaten for their insightful comments. 
We also would like to thank Alexander M. Rush for his help in preparing the data set for the summarization task and Sam Gross for providing the image features. 

\bibliography{iclr2016_conference}

\begin{thebibliography}{33}
\providecommand{\natexlab}[1]{#1}
\providecommand{\url}[1]{\texttt{#1}}
\expandafter\ifx\csname urlstyle\endcsname\relax
  \providecommand{\doi}[1]{doi: #1}\else
  \providecommand{\doi}{doi: \begingroup \urlstyle{rm}\Url}\fi

\bibitem[Auli \& Gao(2014)Auli and Gao]{auli2014}
Auli, M. and Gao, J.
\newblock Decoder integration and expected bleu training for recurrent neural
  network language models.
\newblock In \emph{Proc. of ACL}, June 2014.

\bibitem[Ba et~al.()Ba, Mnih, and Kavukcuoglu]{ba_iclr15}
Ba, J.L., Mnih, V., and Kavukcuoglu, K.
\newblock Multiple object recognition with visual attention.

\bibitem[Bahdanau et~al.(2015)Bahdanau, Cho, and Bengio]{bahdanau-iclr2015}
Bahdanau, D., Cho, K., and Bengio, Y.
\newblock Neural machine translation by jointly learning to align and
  translate.
\newblock In \emph{ICLR}, 2015.

\bibitem[Bengio et~al.(2015)Bengio, Vinyals, Jaitly, and
  Shazeer]{sbengio-nips2015}
Bengio, S., Vinyals, O., Jaitly, N., and Shazeer, N.
\newblock Scheduled sampling for sequence prediction with recurrent neural
  networks.
\newblock In \emph{NIPS}, 2015.

\bibitem[Cettolo et~al.(2014)Cettolo, Niehues, St{\"u}ker, Bentivogli, , and
  Federico]{cettolo2014}
Cettolo, M., Niehues, J., St{\"u}ker, S., Bentivogli, L., , and Federico, M.
\newblock Report on the 11th iwslt evaluation campaign.
\newblock In \emph{Proc. of IWSLT}, 2014.

\bibitem[Daume~III et~al.(2009)Daume~III, Langford, and Marcu]{searn}
Daume~III, H., Langford, J., and Marcu, D.
\newblock Search-based structured prediction as classification.
\newblock \emph{Machine Learning Journal}, 2009.

\bibitem[Deng et~al.(2009)Deng, Dong, Socher, Li, Li, and
  Fei-Fei]{imagenet_cvpr09}
Deng, J., Dong, W., Socher, R., Li, L.J., Li, K., and Fei-Fei, L.
\newblock Imagenet: a large-scale hierarchical image database.
\newblock In \emph{IEEE Conference on Computer Vision and Pattern Recognition},
  2009.

\bibitem[Elman(1990)]{elman1990}
Elman, Jeffrey~L.
\newblock Finding structure in time.
\newblock \emph{Cognitive Science}, 14\penalty0 (2):\penalty0 179--211, 1990.

\bibitem[Graff et~al.(2003)Graff, Kong, Chen, and Maeda]{gigaword}
Graff, D., Kong, J., Chen, K., and Maeda, K.
\newblock English gigaword.
\newblock Technical report, 2003.

\bibitem[Graves \& Jaitly(2014)Graves and Jaitly]{graves_icml14}
Graves, A. and Jaitly, N.
\newblock Towards end-to-end speech recognition with recurrent neural networks.
\newblock In \emph{ICML}, 2014.

\bibitem[He \& Deng(2012)He and Deng]{he12}
He, X. and Deng, L.
\newblock Maximum expected bleu training of phrase and lexicon translation
  models.
\newblock In \emph{ACL}, 2012.

\bibitem[Hochreiter \& Schmidhuber(1997)Hochreiter and Schmidhuber]{lstm}
Hochreiter, S. and Schmidhuber, J.
\newblock Long short-term memory.
\newblock \emph{Neural Computation}, 9\penalty0 (8):\penalty0 1735--1780, 1997.

\bibitem[Kneser \& Ney(1995)Kneser and Ney]{kneser+ney1995}
Kneser, Reinhard and Ney, Hermann.
\newblock Improved backing-off for {M}-gram language modeling.
\newblock In \emph{Proc. of the International Conference on Acoustics, Speech,
  and Signal Processing}, pp.\  181--184, May 1995.

\bibitem[Koehn et~al.(2007)Koehn, Hoang, Birch, Callison-Burch, Federico,
  Bertoldi, Cowan, Shen, Moran, Zens, Dyer, Bojar, Constantin, and
  Herbst]{koehn2007}
Koehn, Philipp, Hoang, Hieu, Birch, Alexandra, Callison-Burch, Chris, Federico,
  Marcello, Bertoldi, Nicola, Cowan, Brooke, Shen, Wade, Moran, Christine,
  Zens, Richard, Dyer, Chris, Bojar, Ondrej, Constantin, Alexandra, and Herbst,
  Evan.
\newblock Moses: Open source toolkit for statistical machine translation.
\newblock In \emph{Proc. of ACL Demo and Poster Sessions}, Jun 2007.

\bibitem[Liang et~al.(2006)Liang, Bouchard-C{\^o}t{\'e}, Taskar, and
  Klein]{liang2006}
Liang, Percy, Bouchard-C{\^o}t{\'e}, Alexandre, Taskar, Ben, and Klein, Dan.
\newblock An end-to-end discriminative approach to machine translation.
\newblock In \emph{acl-coling2006}, pp.\  761--768, Jul 2006.

\bibitem[Lin \& Hovy(2003)Lin and Hovy]{rouge}
Lin, C.Y. and Hovy, E.H.
\newblock Automatic evaluation of summaries using n-gram co-occurrence
  statistics.
\newblock In \emph{HLT-NAACL}, 2003.

\bibitem[Lin et~al.(2014)Lin, Maire, Belongie, Hays, Perona, Ramanan, Dollar,
  and Zitnick]{mscoco}
Lin, T.Y., Maire, M., Belongie, S., Hays, J., Perona, P., Ramanan, D., Dollar,
  P., and Zitnick, C.L.
\newblock Microsoft coco: Common objects in context.
\newblock Technical report, 2014.

\bibitem[McAllester et~al.(2010)McAllester, Hazan, and Keshet]{mcallister2010}
McAllester, D., Hazan, T., and Keshet, J.
\newblock Direct loss minimization for structured prediction.
\newblock In \emph{NIPS}, 2010.

\bibitem[Mikolov et~al.(2010)Mikolov, Karafiát, Burget, Cernocký, and
  Khudanpur]{mikolov-2010}
Mikolov, T., Karafiát, M., Burget, L., Cernocký, J., and Khudanpur, S.
\newblock Recurrent neural network based language model.
\newblock In \emph{INTERSPEECH}, 2010.

\bibitem[Mnih et~al.(2014)Mnih, Graves, and Kavukcuoglu]{vmnih-nips2014}
Mnih, V., Heess~N., Graves, A., and Kavukcuoglu, K.
\newblock Recurrent models of visual attention.
\newblock In \emph{NIPS}, 2014.

\bibitem[Morin \& Bengio(2005)Morin and Bengio]{nlm}
Morin, F. and Bengio, Y.
\newblock Hierarchical probabilistic neural network language model.
\newblock In \emph{AISTATS}, 2005.

\bibitem[Papineni et~al.(2002)Papineni, Roukos, Ward, and Zhu]{bleu}
Papineni, K., Roukos, S., Ward, T., and Zhu, W.J.
\newblock Bleu: a method for automatic evaluation of machine translation.
\newblock In \emph{ACL}, 2002.

\bibitem[Reiter \& Dale(2000)Reiter and Dale]{textgen}
Reiter, E. and Dale, R.
\newblock \emph{Building natural language generation systems}.
\newblock Cambridge university press, 2000.

\bibitem[Ross et~al.(2011)Ross, Gordon, and Bagnell]{dagger}
Ross, S., Gordon, G.J., and Bagnell, J.A.
\newblock A reduction of imitation learning and structured prediction to
  no-regret online learning.
\newblock In \emph{AISTATS}, 2011.

\bibitem[Rosti et~al.(2011)Rosti, Zhang, Matsoukas, and Schwartz]{rosti2011}
Rosti, Antti-Veikko~I, Zhang, Bing, Matsoukas, Spyros, and Schwartz, Richard.
\newblock Expected bleu training for graphs: Bbn system description for wmt11
  system combination task.
\newblock In \emph{Proc. of WMT}, pp.\  159--165. Association for Computational
  Linguistics, July 2011.

\bibitem[Rumelhart et~al.(1986)Rumelhart, Hinton, and Williams]{bptt}
Rumelhart, D.E., Hinton, G.E., and Williams, R.J.
\newblock Learning internal representations by back-propagating errors.
\newblock \emph{Nature}, 323:\penalty0 533--536, 1986.

\bibitem[Rush et~al.(2015)Rush, Chopra, and Weston]{rush-2015}
Rush, A.M., Chopra, S., and Weston, J.
\newblock A neural attention model for abstractive sentence summarization.
\newblock In \emph{EMNLP}, 2015.

\bibitem[Sutskever et~al.(2014)Sutskever, Vinyals, and Le]{sutskever2014}
Sutskever, Ilya, Vinyals, Oriol, and Le, Quoc.
\newblock Sequence to sequence learning with neural networks.
\newblock In \emph{Proc. of NIPS}, 2014.

\bibitem[Sutton \& Barto(1988)Sutton and Barto]{sutton-rl}
Sutton, R.S. and Barto, A.G.
\newblock \emph{Reinforcement learning: An introduction}.
\newblock MIT Press, 1988.

\bibitem[Venkatraman et~al.(2015)Venkatraman, Hebert, and Bagnell]{dad}
Venkatraman, A., Hebert, M., and Bagnell, J.A.
\newblock Improving multi-step prediction of learned time series models.
\newblock In \emph{AAAI}, 2015.

\bibitem[Williams(1992)]{reinforce}
Williams, R.~J.
\newblock Simple statistical gradient-following algorithms for connectionist
  reinforcement learning.
\newblock \emph{Machine Learning}, 8:\penalty0 229–--256, 1992.

\bibitem[Xu et~al.(2015)Xu, Ba, Kiros, Courville, Salakhutdinov, Zemel, and
  Bengio]{xu-icml2015}
Xu, X., Ba, J., Kiros, R., Courville, A., Salakhutdinov, R., Zemel, R., and
  Bengio, Y.
\newblock Show, attend and tell: Neural image caption generation with visual
  attention.
\newblock In \emph{ICML}, 2015.

\bibitem[Zaremba \& Sutskever(2015)Zaremba and Sutskever]{zaremba-arxiv2015}
Zaremba, W. and Sutskever, I.
\newblock Reinforcement learning neural turing machines.
\newblock Technical report, 2015.

\end{thebibliography}
\bibliographystyle{iclr2016_conference}
\newpage
\section{Supplementary Material}

\subsection{Experiments}

\subsubsection{Qualitative Comparison}
\begin{figure}[h!]
\tiny{
\begin{verbatim}


CONTEXT:      a chinese government official on sunday dismissed reports that the government was delaying the issuing
              of third generation -lrb- #g -rrb- mobile phone licenses in order to give a developing <unk> system an
              advantage
GROUND TRUTH: foreign phone operators to get equal access to china 's #g market
XENT:         china dismisses report of #g mobile phone phone
DAD:          china denies <unk> <unk> mobile phone licenses
E2E:          china 's mobile phone licenses delayed
MIXER:        china official dismisses reports of #g mobile licenses

CONTEXT:      greece risks bankruptcy if it does not take radical extra measures to fix its finances , prime minister
              george papandreou warned on tuesday , saying the country was in a `` wartime situation
GROUND TRUTH: greece risks bankruptcy without radical action
XENT:         greece warns <unk> measures to <unk> finances
DAD:          greece says no measures to <unk> <unk>
E2E:          greece threatens to <unk> measures to <unk> finances
MIXER:        greece does not take radical measures to <unk> deficit

CONTEXT:      the indonesian police were close to identify the body parts resulted from the deadly explosion in front
              of the australian embassy by the dna test , police chief general <unk> <unk> said on wednesday
GROUND TRUTH: indonesian police close to <unk> australian embassy bomber
XENT:         indonesian police close to <unk>
DAD:          indonesian police close to <unk>
E2E:          indonesian police close to monitor deadly australia
MIXER:        indonesian police close to <unk> parts of australian embassy

CONTEXT:      hundreds of catholic and protestant youths attacked security forces with <unk> bombs in a flashpoint
              area of north belfast late thursday as violence erupted for the second night in a row , police said
GROUND TRUTH: second night of violence erupts in north belfast
XENT:         urgent hundreds of catholic and <unk> <unk> in <unk>
DAD:          hundreds of belfast <unk> <unk> in n. belfast
E2E:          hundreds of catholic protestant , <unk> clash with <unk>
MIXER:        hundreds of catholic <unk> attacked in north belfast

CONTEXT:      uganda 's lord 's resistance army -lrb- lra -rrb- rebel leader joseph <unk> is planning to join his
              commanders in the ceasefire area ahead of talks with the government , ugandan army has said
GROUND TRUTH: rebel leader to move to ceasefire area         
XENT:         uganda 's <unk> rebel leader to join ceasefire
DAD:          ugandan rebel leader to join ceasefire talks
E2E:          ugandan rebels <unk> rebel leader
MIXER:        ugandan rebels to join ceasefire in <unk>

CONTEXT:      a russian veterinary official reported a fourth outbreak of dead domestic poultry in a suburban
              moscow district sunday as experts tightened <unk> following confirmation of the presence of the 
              deadly h#n# bird flu strain
GROUND TRUTH: tests confirm h#n# bird flu strain in # <unk> moscow <unk>
XENT:         russian official reports fourth flu in <unk>
DAD:          bird flu outbreak in central china
E2E:          russian official official says outbreak outbreak outbreak in <unk>
MIXER:        russian official reports fourth bird flu

CONTEXT:      a jewish human rights group announced monday that it will offer <unk> a dlrs ##,### reward for 
              information that helps them track down those suspected of participating in nazi atrocities during 
              world war ii
GROUND TRUTH: jewish human rights group offers reward for information on nazi suspects in lithuania
XENT:         jewish rights group announces <unk> to reward for war during world war
DAD:          rights group announces <unk> dlrs dlrs dlrs reward
E2E:          jewish rights group offers reward for <unk>
MIXER:        jewish human rights group to offer reward for <unk>

CONTEXT:      a senior u.s. envoy reassured australia 's opposition labor party on saturday that no decision 
              had been made to take military action against iraq and so no military assistance had been sought 
              from australia
GROUND TRUTH: u.s. envoy meets opposition labor party to discuss iraq
XENT:         australian opposition party makes progress on military action against iraq
DAD:          australian opposition party says no military action against iraq
E2E:          us envoy says no decision to take australia 's labor
MIXER:        u.s. envoy says no decision to military action against iraq

CONTEXT:      republican u.s. presidential candidate rudy giuliani met privately wednesday with iraqi president 
              jalal talabani and indicated that he would keep a u.s. presence in iraq for as long as necessary , 
              campaign aides said
GROUND TRUTH: giuliani meets with iraqi president , discusses war
XENT:         <unk> meets with president of iraqi president
DAD:          republican presidential candidate meets iraqi president
E2E:          u.s. president meets with iraqi president
MIXER:        u.s. presidential candidate giuliani meets with iraqi president
\end{verbatim}
}
\caption{Examples of greedy generations after conditioning on sentences from the test summarization dataset. The "$<$unk$>$" token is produced by our tokenizer and it replaces rare words.} 
\label{fig:generation}
\end{figure}

\subsubsection{Hyperparameters}
\begin{table}[!h]
\caption{Best scheduling parameters found by hyper-parameter search of MIXER.}
\begin{tabular}{l || l | l | l}
      \multicolumn{1}{c||}{\emph{TASK} }  & 
      \multicolumn{1}{c|}{$N^{\mbox{\small XENT}}$} & 
      \multicolumn{1}{c|}{$N^{\mbox{\small XE+R}}$} & \multicolumn{1}{c}{$\Delta$} \\
      \hline
      \hline
      {\em summarization} & 20 & 5 & 2  \\
      \hline
      {\em machine translation} & 25 & 5 & 3 \\
      \hline
      {\em image captioning} & 20 & 5 & 2  \\
    \end{tabular}
\label{tab:scheduling}
\end{table}

\subsubsection{Relative Gains}

\begin{figure}[!h]
\begin{center}
 \includegraphics[width=0.6\linewidth]{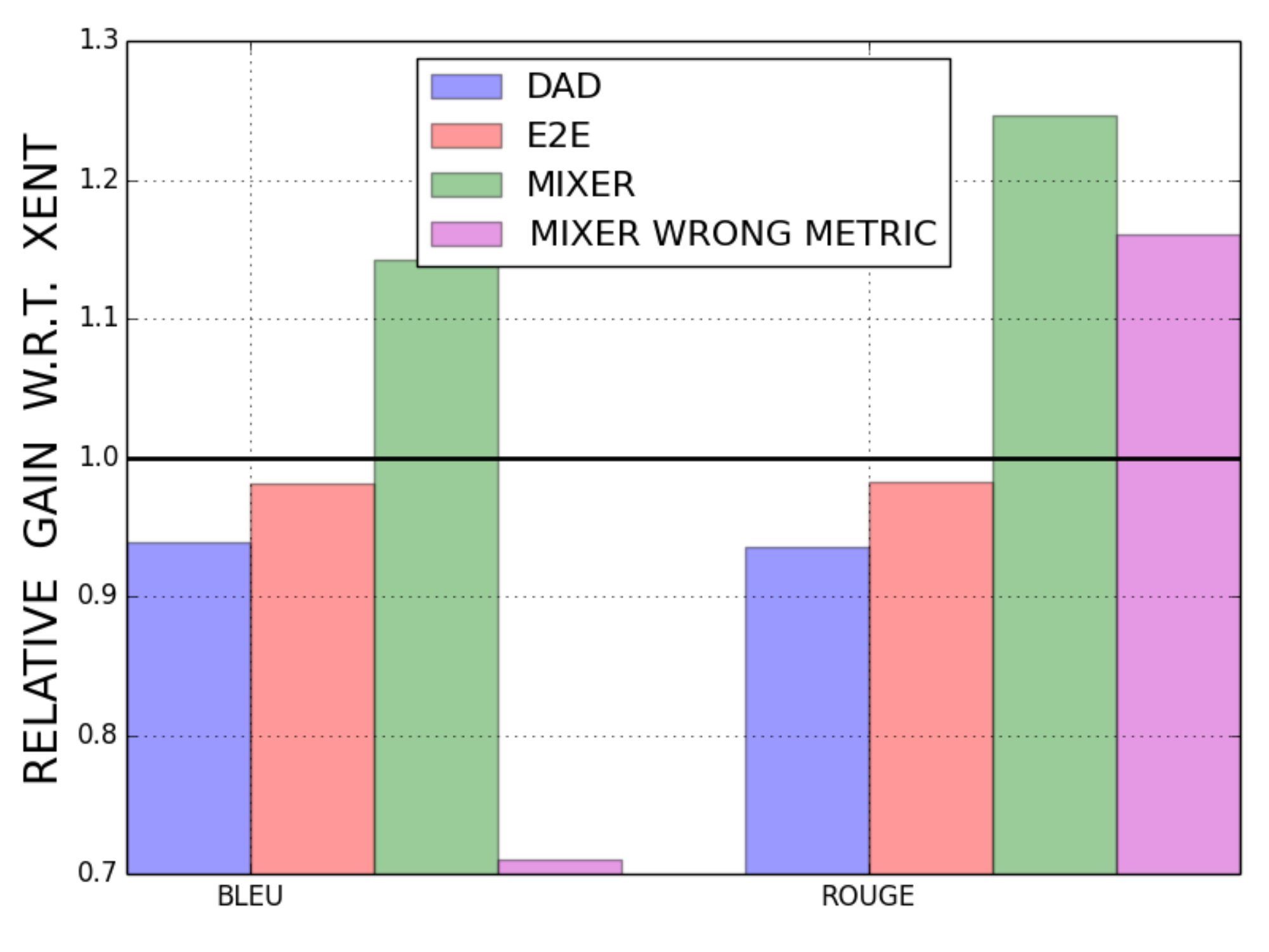}
 \end{center}
\caption{Relative gains on summarization with respect to the XENT baseline. Left: relative BLEU score. Right: relative ROUGE-2.
The models are: DAD, E2E, MIXER trained for the objective used at test time (method proposed in this paper), and MIXER trained with a different metric.
When evaluating for BLEU, the last column on the left reports the evaluation of MIXER trained using ROUGE-2.
When evaluating for ROUGE-2, the last column on the right reports the evaluation of MIXER trained using BLEU.}
\label{fig:summarization_bleu_rouge}
\end{figure}

\vspace{0.3in}
\begin{figure}[!h]
  \centering
  \begin{minipage}{0.05\linewidth}
  	\includegraphics[width=2.5\linewidth]{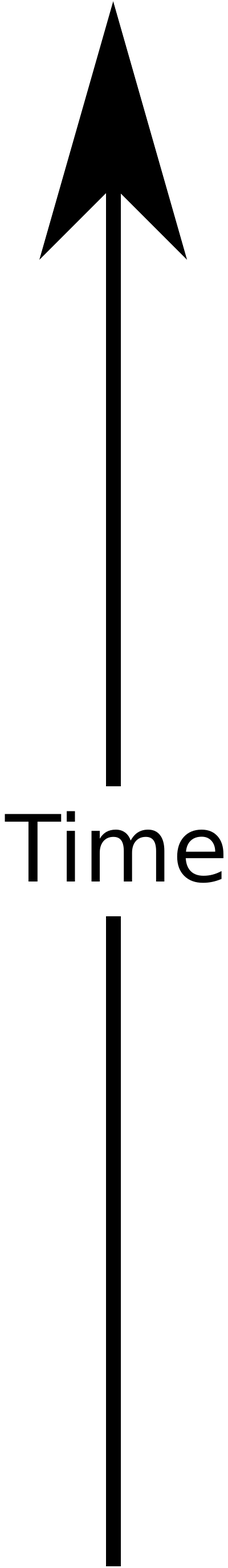}\\
  \end{minipage}
  \begin{minipage}{0.46\linewidth}
  	\centering
    \scalebox{0.75}{
    \tiny
    \begingroup\makeatletter\def\f@size{1}\check@mathfonts
    \begin{tikzpicture}
      \node [draw, circle] (w) at (0,0) {$w$};
      \node [draw, circle] (w0) at (-0.5,1) {$w_0$};
      \node [draw, circle] (w1) at (0.5,1) {$w_1$};
      \node [draw, circle] (w00) at (-1.5,2) {$w_{00}$};
      \node [draw, circle] (w01) at (-0.5,2) {$w_{01}$};
      \node [draw, circle] (w10) at (0.5,2) {$w_{10}$};
      \node [draw, circle] (w11) at (1.5,2) {$w_{11}$};
      \node [draw, circle] (w000) at (-3.5,3) {$w_{000}$};
      \node [draw, circle] (w001) at (-2.5,3) {$w_{001}$};
      \node [draw, circle] (w010) at (-1.5,3) {$w_{010}$};
      \node [draw, circle] (w011) at (-0.5,3) {$w_{011}$};
      \node [draw, circle] (w100) at (0.5,3) {$w_{100}$};
      \node [draw, circle] (w101) at (1.5,3) {$w_{101}$};
      \node [draw, circle] (w110) at (2.5,3) {$w_{110}$};
      \node [draw, circle] (w111) at (3.5,3) {$w_{111}$};
      \node [draw, circle] (w0000) at (-3.75,4) {$w_{\dots}$};
      \node [draw, circle] (w0001) at (-3.25,4) {$w_{\dots}$};
      \node [draw, circle] (w0010) at (-2.75,4) {$w_{\dots}$};
      \node [draw, circle] (w0011) at (-2.25,4) {$w_{\dots}$};
      \node [draw, circle] (w0100) at (-1.75,4) {$w_{\dots}$};
      \node [draw, circle] (w0101) at (-1.25,4) {$w_{\dots}$};
      \node [draw, circle] (w0110) at (-0.75,4) {$w_{\dots}$};
      \node [draw, circle] (w0111) at (-0.25,4) {$w_{\dots}$};
      \node [draw, circle] (w1000) at (0.25,4) {$w_{\dots}$};
      \node [draw, circle] (w1001) at (0.75,4) {$w_{\dots}$};
      \node [draw, circle] (w1010) at (1.25,4) {$w_{\dots}$};
      \node [draw, circle] (w1011) at (1.75,4) {$w_{\dots}$};
      \node [draw, circle] (w1100) at (2.25,4) {$w_{\dots}$};
      \node [draw, circle] (w1101) at (2.75,4) {$w_{\dots}$};
      \node [draw, circle] (w1110) at (3.25,4) {$w_{\dots}$};
      \node [draw, circle] (w1111) at (3.75,4) {$w_{\dots}$};
      \draw [thick] [->] (w) to (w0);
      \draw [thick] [dashed] [green] [->] (w) to (w1);
      \draw [thick] [dashed] [green] [->] (w0) to (w00);
      \draw [thick] [->] (w0) to (w01);
      \draw [dotted] [->] (w1) to (w10);
      \draw [dotted] [->] (w1) to (w11);
      \draw [dotted] [->] (w00) to (w000);
      \draw [dotted] [->] (w00) to (w001);
      \draw [thick] [dashed] [green] [->] (w01) to (w010);
      \draw [thick] [->] (w01) to (w011);
      \draw [dotted] [->] (w10) to (w100);
      \draw [dotted] [->] (w10) to (w101);
      \draw [dotted] [->] (w11) to (w110);
      \draw [dotted] [->] (w11) to (w111);
      \draw [dotted] [->] (w000) to (w0000);
      \draw [dotted] [->] (w000) to (w0001);
      \draw [dotted] [->] (w001) to (w0010);
      \draw [dotted] [->] (w001) to (w0011);
      \draw [dotted] [->] (w010) to (w0100);
      \draw [dotted] [->] (w010) to (w0101);
      \draw [thick] [->] (w011) to (w0110);
      \draw [thick] [dashed] [green] [->] (w011) to (w0111);
      \draw [dotted] [->] (w100) to (w1000);
      \draw [dotted] [->] (w100) to (w1001);
      \draw [dotted] [->] (w101) to (w1010);
      \draw [dotted] [->] (w101) to (w1011);
      \draw [dotted] [->] (w110) to (w1100);
      \draw [dotted] [->] (w110) to (w1101);
      \draw [dotted] [->] (w111) to (w1110);
      \draw [dotted] [->] (w111) to (w1111);
      \end{tikzpicture}
      \endgroup
    }\\
    Training with exposure bias
  \end{minipage}
  \begin{minipage}{0.46\linewidth}
  	\centering
    \scalebox{0.75}{
    \tiny
    \begingroup\makeatletter\def\f@size{1}\check@mathfonts
    \begin{tikzpicture}
      \node [draw, circle] (w) at (0,0) {$w$};
      \node [draw, circle] (w0) at (-0.5,1) {$w_0$};
      \node [draw, circle] (w1) at (0.5,1) {$w_1$};
      \node [draw, circle] (w00) at (-1.5,2) {$w_{00}$};
      \node [draw, circle] (w01) at (-0.5,2) {$w_{01}$};
      \node [draw, circle] (w10) at (0.5,2) {$w_{10}$};
      \node [draw, circle] (w11) at (1.5,2) {$w_{11}$};
      \node [draw, circle] (w000) at (-3.5,3) {$w_{000}$};
      \node [draw, circle] (w001) at (-2.5,3) {$w_{001}$};
      \node [draw, circle] (w010) at (-1.5,3) {$w_{010}$};
      \node [draw, circle] (w011) at (-0.5,3) {$w_{011}$};
      \node [draw, circle] (w100) at (0.5,3) {$w_{100}$};
      \node [draw, circle] (w101) at (1.5,3) {$w_{101}$};
      \node [draw, circle] (w110) at (2.5,3) {$w_{110}$};
      \node [draw, circle] (w111) at (3.5,3) {$w_{111}$};
      \node [draw, circle] (w0000) at (-3.75,4) {$w_{\dots}$};
      \node [draw, circle] (w0001) at (-3.25,4) {$w_{\dots}$};
      \node [draw, circle] (w0010) at (-2.75,4) {$w_{\dots}$};
      \node [draw, circle] (w0011) at (-2.25,4) {$w_{\dots}$};
      \node [draw, circle] (w0100) at (-1.75,4) {$w_{\dots}$};
      \node [draw, circle] (w0101) at (-1.25,4) {$w_{\dots}$};
      \node [draw, circle] (w0110) at (-0.75,4) {$w_{\dots}$};
      \node [draw, circle] (w0111) at (-0.25,4) {$w_{\dots}$};
      \node [draw, circle] (w1000) at (0.25,4) {$w_{\dots}$};
      \node [draw, circle] (w1001) at (0.75,4) {$w_{\dots}$};
      \node [draw, circle] (w1010) at (1.25,4) {$w_{\dots}$};
      \node [draw, circle] (w1011) at (1.75,4) {$w_{\dots}$};
      \node [draw, circle] (w1100) at (2.25,4) {$w_{\dots}$};
      \node [draw, circle] (w1101) at (2.75,4) {$w_{\dots}$};
      \node [draw, circle] (w1110) at (3.25,4) {$w_{\dots}$};
      \node [draw, circle] (w1111) at (3.75,4) {$w_{\dots}$};

      \draw [thick] [->] (w) to (w0);
      \draw [thick] [blue] [->] (w) to (w1);
      \draw [thick] [dashed] [green] [->] (w0) to (w00);
      \draw [thick] [->] (w0) to (w01);
      \draw [thick] [blue] [->] (w1) to (w10);
      \draw [thick] [dashed] [green] [->] (w1) to (w11);
      \draw [thick] [dashed] [green] [->] (w00) to (w000);
      \draw [thick] [dashed] [green] [->] (w00) to (w001);
      \draw [thick] [dashed] [green] [->] (w01) to (w010);
      \draw [thick] [->] (w01) to (w011);
      \draw [thick] [dashed] [green] [->] (w10) to (w100);
      \draw [thick] [blue] [->] (w10) to (w101);
      \draw [thick] [dashed] [green] [->] (w11) to (w110);
      \node [draw, circle] (w000) at (-3.5,3) {$w_{000}$};
      \node [draw, circle] (w001) at (-2.5,3) {$w_{001}$};
      \node [draw, circle] (w010) at (-1.5,3) {$w_{010}$};
      \node [draw, circle] (w011) at (-0.5,3) {$w_{011}$};
      \node [draw, circle] (w100) at (0.5,3) {$w_{100}$};
      \node [draw, circle] (w101) at (1.5,3) {$w_{101}$};
      \node [draw, circle] (w110) at (2.5,3) {$w_{110}$};
      \node [draw, circle] (w111) at (3.5,3) {$w_{111}$};
      \node [draw, circle] (w0000) at (-3.75,4) {$w_{\dots}$};
      \node [draw, circle] (w0001) at (-3.25,4) {$w_{\dots}$};
      \node [draw, circle] (w0010) at (-2.75,4) {$w_{\dots}$};
      \node [draw, circle] (w0011) at (-2.25,4) {$w_{\dots}$};
      \node [draw, circle] (w0100) at (-1.75,4) {$w_{\dots}$};
      \node [draw, circle] (w0101) at (-1.25,4) {$w_{\dots}$};
      \node [draw, circle] (w0110) at (-0.75,4) {$w_{\dots}$};
      \node [draw, circle] (w0111) at (-0.25,4) {$w_{\dots}$};
      \node [draw, circle] (w1000) at (0.25,4) {$w_{\dots}$};
      \node [draw, circle] (w1001) at (0.75,4) {$w_{\dots}$};
      \node [draw, circle] (w1010) at (1.25,4) {$w_{\dots}$};
      \node [draw, circle] (w1011) at (1.75,4) {$w_{\dots}$};
      \node [draw, circle] (w1100) at (2.25,4) {$w_{\dots}$};
      \node [draw, circle] (w1101) at (2.75,4) {$w_{\dots}$};
      \node [draw, circle] (w1110) at (3.25,4) {$w_{\dots}$};
      \node [draw, circle] (w1111) at (3.75,4) {$w_{\dots}$};
      \draw [thick] [dashed] [green] [->] (w000) to (w0000);
      \draw [thick] [dashed] [green] [->] (w000) to (w0001);
      \draw [thick] [dashed] [green] [->] (w001) to (w0010);
      \draw [thick] [dashed] [green] [->] (w001) to (w0011);
      \draw [thick] [dashed] [green] [->] (w010) to (w0100);
      \draw [thick] [dashed] [green] [->] (w010) to (w0101);
      \draw [thick] [->] (w011) to (w0110);
      \draw [thick] [dashed] [green] [->] (w011) to (w0111);
      \draw [thick] [dashed] [green] [->] (w100) to (w1000);
      \draw [thick] [dashed] [green] [->] (w100) to (w1001);
      \draw [thick] [dashed] [green] [->] (w101) to (w1010);
      \draw [thick] [blue] [->] (w101) to (w1011);
      \draw [thick] [dashed] [green] [->] (w110) to (w1100);
      \draw [thick] [dashed] [green] [->] (w110) to (w1101);
      \draw [thick] [dashed] [green] [->] (w111) to (w1110);
      \draw [thick] [dashed] [green] [->] (w111) to (w1111);
      \end{tikzpicture}
      \endgroup
    }
	Training in expectation (Reinforce)
  \end{minipage}
  \caption{Search space for the toy case of a binary vocabulary and sequences of length 4.  
    The trees represent all the $2^4$ possible sequences.
  The solid black line is the ground truth sequence. 
  \textbf{(Left)} Greedy training such as XENT optimizes only the probability 
  of the next word. The model may consider choices indicated by the green arrows, but it starts off from words taken from the ground truth sequence. The model experiences exposure bias, since it sees only words branching off the ground truth path;
  \textbf{(Right)} REINFORCE and MIXER optimize over all possible sequences, using the predictions made by the model itself.  
  In practice, the model samples only a single path indicated by the blue solid line. The model does not suffer from exposure bias; the model is trained as it is tested.}
  \label{fig:plan}
\end{figure}

\subsection{The Attentive Encoder}
\label{sup-material:encoder}
Here we explain in detail how we generate the conditioning vector $\bc_t$ 
for our RNN using the source sentence and the current hidden state $\bh_t$. 
Let us denote by $\bs$ the source sentence which is composed of a sequence 
of $M$ words $\bs = [w_1, \ldots, w_M]$. 
With a slight overload of notation let $w_i$ also denote the $d$ dimensional 
learnable embedding of the $i$-th word ($w_i \in \setr^d$). In addition the 
position $i$ of the word $w_i$ is also associated with a learnable embedding $l_i$ 
of size $d$ ($l_i \in \setr^d$). Then the full embedding for 
the $i$-th word in the input sentence is given by $a_i = w_i + l_i$. 
In order for the embeddings to capture local context, we associate an 
aggregate embedding $z_i$ to each word in the source sentence. 
In particular for a word in the $i$-th position, its aggregate embedding 
$z_i$ is computed by taking a window of $q$ consecutive words centered 
at position $i$ and averaging the embeddings of all the words in this window. More precisely, the aggregate embedding $z_i$ is given by: 
\begin{equation}
z_i = \frac{1}{q} \sum_{h = -q/2}^{q/2} a_{i + h}. 
\end{equation}
In our experiments the width $q$ was set to $5$. 
In order to account for the words at the two boundaries of the input 
sentence we first pad the sequence on both sides with dummy words before 
computing the aggregate vectors $z_i$s. Given these aggregate vectors of words, we compute 
the context vector $c_t$ (the final output of the encoder) as: 
\begin{equation}
c_t = \sum_{j=1}^M \alpha_{j,t} w_j, 
\end{equation}
where the weights $\alpha_{j, t}$ are computed as 
\begin{equation}
\alpha_{j, t} = \frac{\exp(z_j \cdot \bh_{t})}{\sum_{i=1}^M \exp(z_i \cdot \bh_{t})}.
\end{equation}

\subsection{Beam Search Algorithm}
\label{sup-material:beam_search}
Equation~\ref{eq:greedy_gen} always chooses the highest scoring next word candidate
at each time step. At test time we can reduce the effect of search error 
by pursuing not only one but $k$ next word candidates at each point, which 
is commonly known as {\it beam search}.
While still approximate, this strategy can recover higher scoring sequences 
that are often also better in terms of our final evaluation metric.
The algorithm maintains the $k$ highest scoring partial
sequences, where $k$ is a hyper-parameter.
Setting $k=1$ reduces the algorithm to a greedy left-to-right search 
(Eq.~\eqref{eq:greedy_gen}). 
\begin{algorithm}[!h]
  \KwIn{model $p_{\theta}$, beam size $k$}
  \KwResult{sequence of words $[w_1^g, w_2^g, \dots, w_n^g]$}
  empty heaps $\{\mathcal{H}_t\}_{t = 1, \dots T}$\;
  an empty hidden state vector $\bh_1$\;
  $\mathcal{H}_1.\mbox{push}(1, [[\varnothing], \bh_1])$\;
  \For{$t\leftarrow 1$ \KwTo $T - 1$}{	
    \For{$i\leftarrow 1$ \KwTo $\min(k, \#\mathcal{H}_t)$}{
      $(p, [[w_1, w_2, \dots, w_t], \bh]) \leftarrow \mathcal{H}_t.\mbox{pop}()$\;  
      $\bh' = \phi_\theta(w, \bh)$ \;      
      \For{$w' \leftarrow$ $k$-most likely words $w'$ from $p_\theta(w' | w_t, \bh)$}{
        $p' = p * p_\theta(w' | w, \bh)$\;
        $\mathcal{H}_{t + 1}.\mbox{push}(p', [[w_1, w_2, \dots, w_t, w'], \bh'])$\;
      }
    }
  }
  $(p, [[w_1, w_2, \dots, w_T], \bh]) \leftarrow \mathcal{H}_T.\mbox{pop}()$\;
  \KwOut{$[w_1, w_2, \dots, w_T]$}
 
 \caption{Pseudo-code of beam search with beam size $k$.}
  \label{alg:beam} 
\end{algorithm}

\newpage

\subsection{Notes}
The current version of the paper updates the first version uploaded on arXiv as follows:
\begin{itemize}
\item on the summarization task, we report results using both ROUGE-2 and BLEU to demonstrate that MIXER can work with any metric.
\item on machine translation and image captioning we use LSTM instead of Elman RNN to demonstrate the MIXER can work with any underlying parametric model.
\item BLEU is evaluated using up to 4-grams, and it is computed at the corpus level (except in the image captioning case) as this seems the most common practice
in the summarization and machine translation literature.
\item we have added several references as suggested by our reviewers
\item we have shortened the paper by moving some content to the Supplementary Material. 
\end{itemize}

\end{document}